\documentclass[letterpaper, 10 pt, conference]{ieeeconf}  

\IEEEoverridecommandlockouts                              

\usepackage{graphics} 
\usepackage[caption=false, font=normalsize, labelfont=sf, textfont=sf]{subfig}
\usepackage[none]{hyphenat}
\usepackage{amsfonts,amsmath,amssymb}
\usepackage{physics}
\usepackage{textcomp}
\usepackage{mathtools}
\usepackage{blindtext}
\usepackage{tabularx}
\usepackage{multirow}
\usepackage{tabularx,booktabs}
\usepackage{algorithmic}
\usepackage{lipsum}
\usepackage{cases}
\usepackage[short]{optidef}
\usepackage{url}
\usepackage[normalem]{ulem}
\usepackage[hidelinks]{hyperref}
\usepackage[dvipsnames]{xcolor}
\usepackage[linesnumbered,ruled,vlined]{algorithm2e}
\usepackage{soul}

\newcommand{\cmmnt}[1]{}

\newcommand{\ie}{\textit{i.e.,}}

\usepackage{tikz}
\newcommand\copyrighttext{%
  \footnotesize \textcopyright 2025 IEEE. Personal use of this material is permitted.
  Permission from IEEE must be obtained for all other uses, in any current or future
  media, including reprinting/republishing this material for advertising or promotional
  purposes, creating new collective works, for resale or redistribution to servers or
  lists, or reuse of any copyrighted component of this work in other works.}
\newcommand\copyrightnotice{%
\begin{tikzpicture}[remember picture,overlay]
\node[anchor=north,yshift=-10pt] at (current page.north) 
  {\fbox{\parbox{\dimexpr\textwidth-\fboxsep-\fboxrule\relax}{\copyrighttext}}};
\end{tikzpicture}%
}

\title{\LARGE \bf Kineto-Dynamical Planning and Accurate Execution\\of Minimum-Time Maneuvers on Three-Dimensional Circuits}

\author{Mattia Piccinini$^{1}$, Sebastiano Taddei$^{2, 3}$, Johannes Betz$^{1}$ and Francesco Biral$^{2}$%
\thanks{*This work was partly supported by the European Union Next-GenerationEU (Piano Nazionale di Ripresa e Resilienza (PNRR) - Missione 4 Componente 1, Investimenti 3.4 e 4.1 - Decreto del Ministero dell'Università e della Ricerca n.351 del 09/04/2022) within the Italian National Ph.D. Program in Autonomous Systems (DAuSy).}
\thanks{$^{1}$Mattia Piccinini and Johannes Betz are with the Professorship of Autonomous Vehicle Systems, Technical University of Munich, 85748 Garching, Germany; Munich Institute of Robotics and Machine Intelligence (MIRMI) {\tt\small name.surname@tum.de}}
\thanks{$^{2}$Sebastiano Taddei and Francesco Biral are with the Department of Industrial Engineering, University of Trento, 38123 Trento, Italy {\tt\small name.surname@unitn.it}}
\thanks{$^{3}$Sebastiano Taddei is also with the Department of Electrical and Information Engineering, Politecnico di Bari, 70125 Bari, Italy {\tt\small s.taddei@phd.poliba.it}}
}

\begin{document}
\copyrightnotice
\maketitle
\thispagestyle{empty}
\pagestyle{empty}
\begin{abstract}
Online planning and execution of minimum-time maneuvers on three-dimensional (3D) circuits is an open challenge in autonomous vehicle racing. In this paper, we present an artificial race driver (ARD) to learn the vehicle dynamics, plan and execute minimum-time maneuvers on a 3D track. ARD integrates a novel kineto-dynamical (KD) vehicle model for trajectory planning with economic nonlinear model predictive control (E-NMPC). We use a high-fidelity vehicle simulator (VS) to compare the closed-loop ARD results with a minimum-lap-time optimal control problem (MLT-VS), solved offline with the same VS. Our ARD sets lap times close to the MLT-VS, and the new KD model outperforms a literature benchmark. Finally, we study the vehicle trajectories, to assess the re-planning capabilities of ARD under execution errors.
A video with the main results is available as supplementary material.
\end{abstract}
\section{Introduction} \label{Intro}
Autonomous vehicle racing (AVR) \cite{Betz2022} has been gaining interest in the research community, as it represents a test-bed for high-performance autonomous driving. An open challenge in AVR is online planning and execution of minimum-time maneuvers on three-dimensional (3D) circuits, which require accurate vehicle dynamic models for trajectory planning.
Many real-world tracks, like the Las Vegas Motor Speedway (20\textdegree{} banking) \cite{Rowold2023} or the Eau Rouge corner at Spa (9\textdegree{} slope), cannot be simplified as 2D planes.

The current autonomous agents often struggle to fully exploit the vehicle performance on 3D tracks, due to trajectory planning inaccuracies. A key unsolved goal is to match the closed-loop lap times with those of minimum-lap-time optimal control problems (MLT-OCPs), using high-fidelity vehicle models.
Our contributions are:
\begin{itemize}
	\item A new kineto-dynamical (KD) vehicle model for trajectory planning with E-NMPC, which extends the one in \cite{PiccininiARD2024} for 3D tracks.
	\item The integration of the KD model into an artificial race driver (ARD), which learns the vehicle dynamics, plans and executes online minimum-time maneuvers with a high-fidelity vehicle simulator (VS), on a 3D circuit.
	\item A comparison of the ARD results with an MLT-OCP, named MLT-VS, solved with the high-fidelity VS.
	\item A comparison of our KD model with the point-mass model of \cite{Rowold2023}, in terms of closed-loop performance.
\end{itemize}
\begin{figure}[]
	\centering
	\includegraphics[scale=0.45]{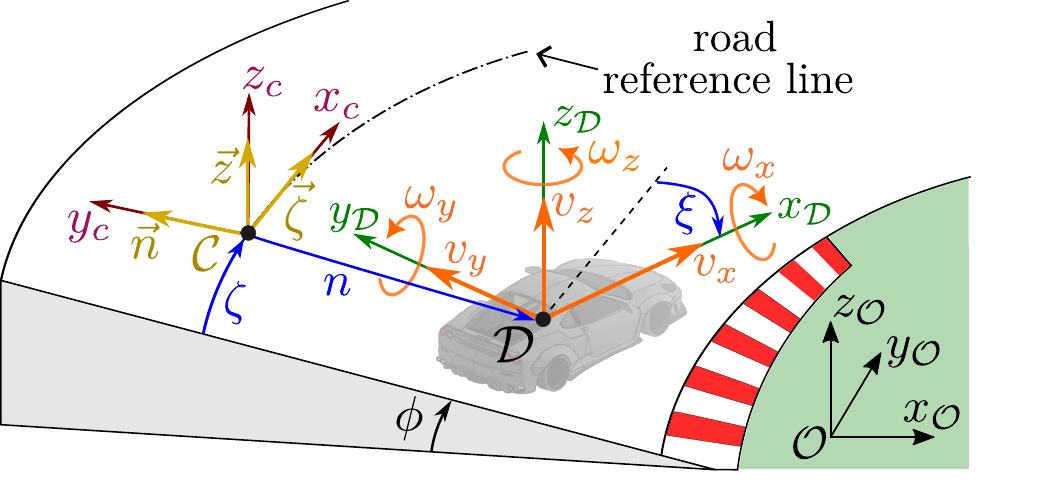}
	\caption{Reference frames and curvilinear coordinates on a 3D road.}
	\label{fig_curvil_coords_3D}
\end{figure}
\subsection{Related Work} \label{sec_related_work}
The authors of \cite{Limebeer2023reviewroad} recently reviewed the 2D and 3D road models for racing vehicle control.
The first MLT-OCPs on 3D roads were solved in \cite{Lot2014,Limebeer2015}, using a ribbon-road model. This model was extended by \cite{Lovato2021,Limebeer2023} and \cite{Fork2024,Fork2021}, for generic non-planar surfaces.
On certain circuits, the road angles were considered \textit{small} \cite{Lot2014,Leonelli2020,Lovato2022,Lovato2022b}, to simplify the equations of motion.

Economic\footnote{The term \textit{economic} MPC is used to disambiguate from tracking MPC.} nonlinear model predictive control (E-NMPC) is becoming popular for trajectory planning in AVR \cite{Betz2022,Piccinini2020,Rowold2023,Piccinini2024_primitives}.
The main challenge for \textit{online} minimum-time planning with E-NMPC on 3D circuits is devising accurate yet computationally efficient models of the vehicle dynamics and performance.
In \cite{Rowold2023,Ogretmen2024}, the maximum performance g-g-v diagram\footnote{The g-g-v diagram (top view in Fig. \ref{fig_ggv_2D_3D_top_Mugello_3D}) is a plot of the maximum lateral and longitudinal vehicle accelerations, as a function of the vehicle speed.} was underestimated, which led to suboptimal lap times. In \cite{Lovato2022,Lovato2022b}, the g-g-v diagrams on 3D roads were estimated with a quasi steady-state (QSS) approach; however, they performed only offline trajectory optimization.
The authors of \cite{subosits2021impacts} solved offline MLT-OCPs on 3D roads, and tracked these maneuvers online using feedback controllers. While their online lap times were close to the offline ones, they tracked fixed solutions without re-planning, and they did not evaluate the trajectory optimality under execution errors.
\subsubsection*{Critical Summary}
To our knowledge, the prior literature is limited by at least one of the following aspects:
\begin{itemize}
	\item Minimum-time trajectory planning on 3D tracks used simplified g-g-v acceleration constraints, which did not accurately capture the performance limits.
	\item The g-g-v diagrams on 3D circuits were estimated with QSS approaches, which assumed the prior knowledge of the vehicle equations of motion.
	\item The performance of online trajectory planning and execution was not compared with MLT-OCPs, solved with high-fidelity vehicle models.
\end{itemize}
\section{Artificial Race Driver on 3D Circuits} \label{sec_framework}
\begin{figure}[h!]
	\centering
	\includegraphics[width=1\columnwidth]{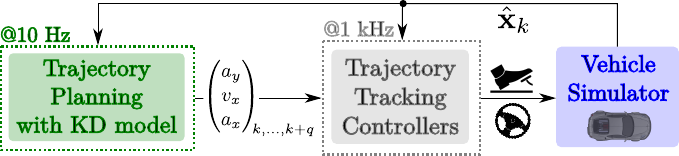}
	\caption{Trajectory planning and control framework of ARD on 3D circuits.}
	\label{fig_planning_control}
\end{figure}

The new KD model of this paper is integrated into the high-level E-NMPC trajectory planner of an artificial race driver (ARD), introduced in \cite{PiccininiARD2024}. Fig. \ref{fig_planning_control} shows the ARD's planning and control framework, with the low-level tracking controllers (gray module) from \cite{Piccinini2023Access}. In this paper, ARD learns to control a high-fidelity vehicle simulator (VS) on a 3D track. Next, we introduce the KD model for trajectory planning, while Section \ref{sec_learning_maximum_performance} discusses the learning process.
%
%
\subsection{Kineto-Dynamical Vehicle Model on a 3D Road} \label{sec_kineto_dynamical_model}
\subsubsection{3D Road Model} \label{sec_road_model}
Following \cite{Lot2014,Limebeer2015}, the road is modeled as a ribbon in $\mathbb{R}^3$. As shown in Fig. \ref{fig_curvil_coords_3D}, we define a moving frame $\mathcal{C}$ that travels along the road reference line\footnote{The road reference line may not coincide with the center-line.}, with unit vectors ${\vec{\boldsymbol{\zeta}}, \vec{\boldsymbol{n}}, \vec{\boldsymbol{z}}}$. Here, $\vec{\boldsymbol{\zeta}}$ points forward along the road, $\vec{\boldsymbol{n}}$ points laterally, and $\vec{\boldsymbol{z}}$ points upward. Starting from the inertial frame $\mathcal{O}$ (Fig. \ref{fig_curvil_coords_3D}), the orientation of the road frame $\mathcal{C}$ is given by three successive rotations:
\begin{equation}
	\label{eq_rot_mat_road_3D}
	\boldsymbol{R}_{\mathcal{C}} = \boldsymbol{R}_z(\theta) \, \boldsymbol{R}_y(-\mu) \, \boldsymbol{R}_x(-\phi) =
	\begin{bmatrix}
		c_{\theta} & -s_{\theta} & -s_{\theta}\phi - c_{\theta} \mu \\
		s_{\theta} & c_{\theta} & c_{\theta} \phi - s_{\theta} \mu \\
		\mu & -\phi & 1
	\end{bmatrix}
\end{equation}
where $\boldsymbol{R}_i$, $i \in \{x,y,z\}$, is the rotation matrix around the axis $i$. The angles $\theta$, $\mu$, and $\phi$ are the road heading, slope, and banking. We define $\mu > 0$ when the road ascends, and $\phi > 0$ when a left turn's banking tilts inward. In \eqref{eq_rot_mat_road_3D}, $c_{\theta}$ and $s_{\theta}$ denote $\cos(\theta)$ and $\sin(\theta)$, respectively.
Assuming small $\mu$ and $\phi$ angles, as done by many authors \cite{Lot2014,Leonelli2020,Lovato2022,Lovato2022b}, is reasonable for the Mugello track used in this paper.

The angles ${\theta,\mu,\phi}$ vary with the road parameter $\zeta$, the curvilinear abscissa of the reference line (Fig. \ref{fig_curvil_coords_3D}). As in \cite{Meirovitch2012}, the change in orientation of the road frame $\mathcal{C}$ is given by:
\begin{equation}
	\label{eq_Wc_skew_sym}
	\boldsymbol{W}_{\mathcal{C}} = \dv{\boldsymbol{R}_{\mathcal{C}}}{\zeta} \, \boldsymbol{R}_{\mathcal{C}}^T
	=
	\begin{bmatrix}
		0 & -\kappa(\zeta) & \upsilon(\zeta) \\
		\kappa(\zeta) & 0 & -\tau(\zeta) \\
		-\upsilon(\zeta) & \tau(\zeta) & 0
	\end{bmatrix}
\end{equation}
Here, $\kappa$, $\upsilon$, and $\tau$ represent the road curvatures in the geodesic ($x_c y_c$), sagittal ($x_c z_c$), and frontal ($y_c z_c$) planes, respectively. Substituting $\boldsymbol{R}_{\mathcal{C}}$ from \eqref{eq_rot_mat_road_3D} into \eqref{eq_Wc_skew_sym}:
\begin{subnumcases}{\label{eq:road_3D_eqns}}
	\theta'(\zeta) = \kappa(\zeta) + \phi(\zeta) \upsilon(\zeta) \label{eq:road_3D_eqns_1} \\
	\mu'(\zeta) = \upsilon(\zeta) - \kappa(\zeta) \phi(\zeta) \label{eq:road_3D_eqns_2} \\
	\phi'(\zeta) = \tau(\zeta) + \kappa(\zeta) \mu(\zeta) \label{eq:road_3D_eqns_3}
\end{subnumcases}
where $\square' = \dv{\square}{\zeta}$.
Thus, the road reference line's orientation is defined by the curvatures $\{\kappa(\zeta), \upsilon(\zeta), \tau(\zeta)\}$ and the initial values $\{\theta(0), \mu(0), \phi(0)\}$.
%
\subsubsection{Vehicle Motion on a 3D Road}
To model the vehicle motion on a 3D road, we introduce a Darboux moving frame\footnote{A Darboux frame is a local moving frame defined on a surface.} $\mathcal{D}$ (Fig. \ref{fig_curvil_coords_3D}), as in \cite{Lot2014}. The origin of $\mathcal{D}$ is the vehicle's center projected onto the road. The $x_{\mathcal{D}}$ axis is at the intersection of the local road tangent plane and the sagittal vehicle plane, while $z_{\mathcal{D}}$ is normal to the road and points upward. The frame $\mathcal{D}$ is defined by the curvilinear coordinates $\{\zeta, n, \xi\}$: $\zeta$ is the curvilinear abscissa, $n$ is the lateral offset, and $\xi$ is the relative yaw angle. The frame moves with linear velocities $\prescript{\mathcal{D}}{}{\boldsymbol{v}} = [v_x, v_y, v_z]$ and angular velocities $\prescript{\mathcal{D}}{}{\boldsymbol{\omega}} = [\omega_x, \omega_y, \omega_z]$, which can be projected in the road frame $\mathcal{C}$ as:
\begin{subnumcases}{\label{eq_curvil_coords}}
	\dot{\zeta} = (v_x \cos(\xi) - v_y \sin(\xi)) \, / \, (1 - n \, \kappa(\zeta)) \label{eq_curvil_coords_1} \\
	\dot{n} = v_x \sin(\xi) + v_y \cos(\xi) \label{eq_curvil_coords_2} \\
	\dot{\xi} = \omega_z - \kappa(\zeta) \, \dot{\zeta} 
	\label{eq_curvil_coords_3} \\
	\omega_x = - \big( \tau\big(\zeta\big) \cos(\xi) + \upsilon(\zeta) \sin(\xi) \big) \, \dot{\zeta} \label{eq_omega_x} \\
	\omega_y = \big(\tau\big(\zeta\big) \sin(\xi) - \upsilon(\zeta) \cos(\xi) \big) \, \dot{\zeta} \label{eq_omega_y} \\
	v_z = - n \, \tau(\zeta) \, \dot{\zeta} \label{eq_vz}
\end{subnumcases}
where $\dot{\square} = \dv{\square}{t}$ indicates the time derivative of $\square$. Finally, the velocity and acceleration vectors $\prescript{\mathcal{D}}{}{\boldsymbol{v}_{\mathrm{c}}}$ and $\prescript{\mathcal{D}}{}{\boldsymbol{a}_{\mathrm{c}}}$ of the vehicle center of mass, whose height from ground is $h$, are:
\begin{subnumcases}{\label{eq_vel_accel_CoM_3D}}
	\prescript{\mathcal{D}}{}{\boldsymbol{v}_{\mathrm{c}}} = \prescript{\mathcal{D}}{}{\boldsymbol{v}} + \prescript{\mathcal{D}}{}{\boldsymbol{\omega}} \cross \begin{bmatrix} 0 & 0 & h \end{bmatrix} \label{eq_vel_CoM_3D} \\
	\prescript{\mathcal{D}}{}{\boldsymbol{a}_{\mathrm{c}}} = \begin{bmatrix} a_{x} & a_{y} & a_{z} \end{bmatrix} = \prescript{\mathcal{D}}{}{\dot{\boldsymbol{v}}_{\mathrm{c}}} + \prescript{\mathcal{D}}{}{\boldsymbol{\omega}} \cross \prescript{\mathcal{D}}{}{\boldsymbol{v}}_{\mathrm{c}} + \prescript{\mathcal{D}}{}{\boldsymbol{g}} \label{eq_accel_CoM_3D}
\end{subnumcases}
where $\boldsymbol{g}$ is the gravitational acceleration vector. \eqref{eq_accel_CoM_3D} yields:
\begin{subnumcases}{\label{eq_accel_CoM_3D_expanded}}
	\begin{aligned}
		a_{x} = \dot{v}_x &- \omega_z (v_y - \omega_x h) + \omega_y v_z + \dot{\omega}_y h + \\ &- g \big( s_{\xi} \phi - c_{\xi} \mu \big) \label{eq_ax_CoM_3D}
	\end{aligned}\\
	\begin{aligned}
		a_{y} = \dot{v}_y &+ \omega_z (v_x + \omega_y h) - \omega_x v_z - \dot{\omega}_x h +\\ &- g \big( s_{\xi} \mu + c_{\xi} \phi \big) \label{eq_ay_CoM_3D}
	\end{aligned}\\
	a_{z} = \dot{v}_z - \omega_y v_x 
	\label{eq_az_CoM_3D}
\end{subnumcases}
\subsubsection{Kineto-Dynamical Model} \label{sec_vehicle_dynamics}
This section presents the kineto-dynamical (KD) vehicle model, extending \cite{Piccinini2023Access} for 3D roads. The model's time-domain equations are:
\begin{subnumcases}{\label{eq_MPC_mod_3D}}
	\tau_{\omega}(v_x) \, \dot{\omega}_z + \omega_z = \omega_{z_0} \, \omega_{z_\mathrm{M}}^{\mathrm{3D}}(v_x,a_z,\xi,\mu,\phi) \label{eq_MPC_mod_3D_Omega} \\
	\tau_{v}(v_x) \, \dot{v}_y + v_y = F_{v}^{\mathrm{3D}}(a_y,v_x,a_x,a_z) \hspace{0.6cm} \label{eq_MPC_mod_3D_vy} \\
	\dot{v}_x = a_x \label{eq_MPC_mod_3D_vx} \\
	\tau_{a} \dot{a}_x + a_x = a_{x_0} \label{eq_MPC_mod_3D_ax} \\
	\eqref{eq_curvil_coords_1}, \eqref{eq_curvil_coords_2}, \eqref{eq_curvil_coords_3} \notag
\end{subnumcases}
%
%
where the state vector is $\boldsymbol{x} = [\omega_z, v_y, v_x, a_x, \zeta, n, \xi]$, and the controls are $\boldsymbol{u} = [\omega_{z_0},a_{x_0}]$.
\subsubsection*{Yaw Rate Dynamics}
Equation \eqref{eq_MPC_mod_3D_Omega} models yaw rate dynamics as a first-order system. The time constant $\tau_{\omega}(v_x)$, an identified second-order polynomial, captures the dynamic dependency on $v_x$. The normalized control $\omega_{z_0} \in [-1,1]$ is scaled by the maximum yaw rate $\omega_{z_\mathrm{M}}^{\mathrm{3D}}(\cdot)$, defined as:
\begin{subnumcases}{\label{eq_omega_zM_3D}}
	a_{y_{\mathrm{M}}}^{\mathrm{2D}}(v_x) = \textstyle \sum_{i=0}^{p_{a_y}} a_{y_{\mathrm{M}_i}} v_x^i \label{eq_ayM_fcn_2D} \\
	S(a_z) = 1 + s_1 \, a_z + s_2 \, a_z^2 \label{eq_S_az} \\
	a_{y_{\mathrm{M}}}^{\mathrm{3D}}(v_x,a_z,\xi,\mu,\phi) = a_{y_{\mathrm{M}}}^{\mathrm{2D}}(v_x) \, S(a_z) + g \, (s_{\xi} \mu + c_{\xi} \phi) \label{eq_ayM_fcn_3D} \\
	\omega_{z_{\mathrm{M}}}^{\mathrm{3D}}(v_x,a_z,\xi,\mu,\phi) = a_{y_{\mathrm{M}}}^{\mathrm{3D}}(v_x,a_z,\xi,\mu,\phi) \, / \, v_x \label{eq_omega_zM_fcn_3D}
\end{subnumcases}
On a 2D road, the maximum lateral acceleration is $a_{y_{\mathrm{M}}}^{\mathrm{2D}}(v_x)$ in \eqref{eq_ayM_fcn_2D}, and is a fitted polynomial in $v_x$. On a 3D road, $a_{y_{\mathrm{M}}}^{\mathrm{3D}}(\cdot)$ also accounts for the gravity, with the term $g \, (s_{\xi} \mu + c_{\xi} \phi)$, and for the vertical acceleration $a_z$, which affects the tire loads, the load transfers, and the lateral performance. The function $S(a_z)$ in \eqref{eq_S_az} is a learnable polynomial that models the dependence of $a_{y_{\mathrm{M}}}^{\mathrm{3D}}(\cdot)$ on $a_z$, as detailed in Section \ref{sec_learning_maximum_performance}. Section \ref{sec_negligible_terms} will show that the terms involving $\omega_y$ and $\omega_x$ in \eqref{eq_ay_CoM_3D} are negligible for $a_{y_{\mathrm{M}}}^{\mathrm{3D}}(\cdot)$, at least for the racetrack considered in this paper. Lastly, the maximum yaw rate $\omega_{z_{\mathrm{M}}}^{\mathrm{3D}}(\cdot)$ in \eqref{eq_omega_zM_fcn_3D} is given by $a_{y_{\mathrm{M}}}^{\mathrm{3D}}(\cdot) / v_x$.
\subsubsection*{Lateral Velocity Dynamics}
Equation \eqref{eq_MPC_mod_3D_vy} is a first-order model of the lateral velocity dynamics. This model is derived from \cite{Piccinini2023Access}, but the function $F_{v}^{\mathrm{3D}}(\cdot)$ now considers the vertical acceleration $a_z$:
\begin{equation}
	\label{eq_Fv_3D}
	\begin{aligned}
		F_{v}^{\mathrm{3D}}(a_y,v_x,a_x,&a_z) = \textstyle \sum_{i=1}^{p_y} \big( \textstyle \sum_{j=0}^{p_v} q_{j_i} v_x^j \big) \, a_y^{2i-1} \cdot \\ & \cdot \big( 1 + \textstyle \sum_{j=1}^{p_x} b_{j_i} a_x^j \big) \, \big( 1 + \textstyle \sum_{j=1}^{p_z} c_{j_i} a_z^j \big)
	\end{aligned}
\end{equation}
where $a_y = \omega_z v_x$. The model \eqref{eq_Fv_3D} can be seen as a (shallow) polynomial neural network \cite{Chrysos2022}, in which the parameters $\{q_{j_i},b_{j_i},c_{j_i}\}$ are learnable. The polynomial degrees are set to $p_y = 3$, $p_v = 4$, $p_x = 2$, and $p_z = 2$ via grid search. The time constant $\tau_{v}(v_x)$ in \eqref{eq_MPC_mod_3D_vy} is an identified polynomial that captures the dynamics' dependency on $v_x$.
\subsubsection*{Longitudinal Dynamics}
The longitudinal velocity $v_x$ dynamics are given by the simple relation \eqref{eq_MPC_mod_3D_vx}, while the bounds on the acceleration $a_x$ consider the 3D road effects (next subsection).
Finally, equation \eqref{eq_MPC_mod_3D_ax} is a first-order relation between the longitudinal acceleration control $a_{x_0}$ and the state $a_x$, with a small time constant $\tau_{a}$.
%
\subsubsection{g-g-v Constraints} \label{sec_ggv_constraints}
\begin{figure}[]
	\centering
	\includegraphics[width=0.80\columnwidth]{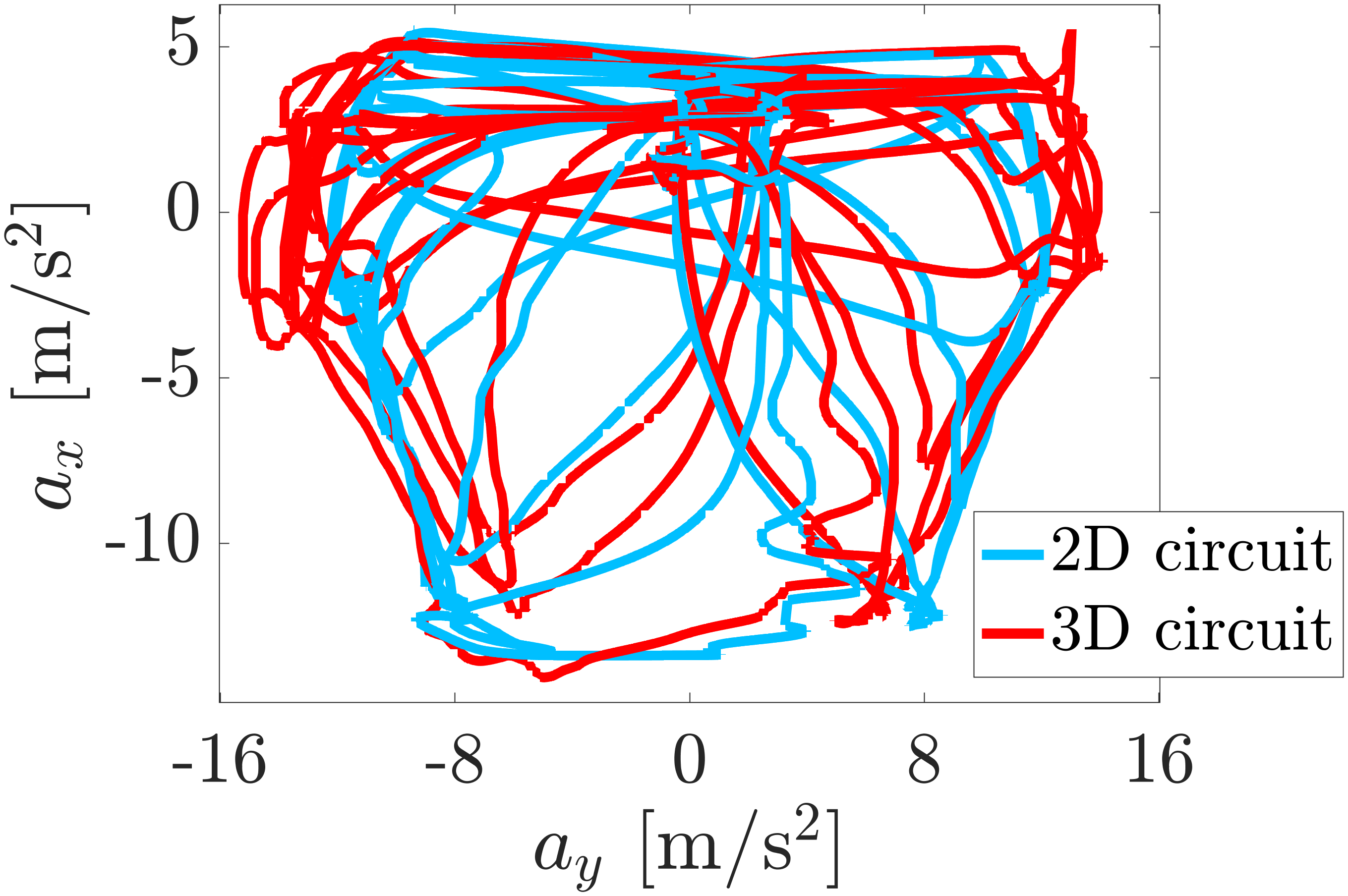}
	\caption{g-g-v diagram (top view) obtained by ARD with closed-loop control of the vehicle simulator, on the 2D and 3D versions of the Mugello circuit. Note the increased lateral accelerations on the 3D track.}
	\label{fig_ggv_2D_3D_top_Mugello_3D}
\end{figure}

Let us now formulate the g-g-v diagram constraints, \ie{} the limits on the lateral and longitudinal accelerations, as a function of $v_x$. Fig. \ref{fig_ggv_2D_3D_top_Mugello_3D} shows a top view of the g-g-v diagram, for the vehicle simulator of this paper, on the 2D and 3D versions of the Mugello circuit. We extend the formulation in \cite{Piccinini2024ggv} to 3D roads:
\begin{subnumcases}{\label{eqn_proposed_ggv}}
	a_{x_{\mathrm{m}}}^{\mathrm{2D}}(v_x) = \textstyle \sum_{i=0}^{p_{\mathrm{m}}} a_{\mathrm{m}_i} v_x^i, \hspace{0.3cm} a_{x_{\mathrm{M}}}^{\mathrm{2D}}(v_x) = \textstyle \sum_{i=0}^{p_{\mathrm{M}}} a_{\mathrm{M}_i} v_x^i \label{eq_ax_min_max_2D} \\
	a_{x_{\mathrm{m}}}^{\mathrm{3D}}(v_x,\xi,\mu,\phi) = a_{x_{\mathrm{m}}}^{\mathrm{2D}}(v_x) + g (s_{\xi} \phi - c_{\xi} \mu ) \label{eq_ax_min} \\
	a_{x_{\mathrm{M}}}^{\mathrm{3D}}(v_x,\xi,\mu,\phi) = a_{x_{\mathrm{M}}}^{\mathrm{2D}}(v_x) + g (s_{\xi} \phi - c_{\xi} \mu ) \label{eq_ax_max} \\
	a_{x_{\mathrm{m}}}^{\mathrm{3D}}(v_x,\xi,\mu,\phi) \leq a_{x} \leq a_{x_{\mathrm{M}}}^{\mathrm{3D}}(v_x,\xi,\mu,\phi) \label{eq_ax_limits} \\
	\boldsymbol{P}
	\begin{bmatrix}
		\frac{a_y - g (s_{\xi} \mu + c_{\xi} \phi)}{S(a_z)} & a_x - g (s_{\xi} \phi - c_{\xi} \mu ) & v_x
	\end{bmatrix}^T \leq \boldsymbol{r}
	\label{eq_ggv_polytopic}\\
	a_x - g (s_{\xi} \phi - c_{\xi} \mu) \geq \bar{s} \, \Big(\Big|\frac{a_y - g (s_{\xi} \mu + c_{\xi} \phi)}{S(a_z)}\Big| - \bar{a}_{y}\Big) \label{eq_ggv_comb_brake}
\end{subnumcases}
where \eqref{eq_ax_min_max_2D} models the minimum and maximum $a_x$ on a 2D road, with identified polynomials in $v_x$. Equations \eqref{eq_ax_min}, \eqref{eq_ax_max}, \eqref{eq_ax_limits} extend the $a_x$ constraints for a 3D road, by adding the gravity terms due to the slope $\mu$ and banking $\phi$ angles.
\eqref{eq_ggv_polytopic} is a set of $\bar{n}$ polytopic constraints, where $\boldsymbol{P} \in \mathbb{R}^{\bar{n} \mathrm{x} 3}$ and $\boldsymbol{r} \in \mathbb{R}^{\bar{n} \mathrm{x} 1}$ are fitted to the identified g-g-v diagram on a 2D road. When the road is 3D, the g-g-v polytope is translated by the gravity terms appearing in \eqref{eq_ggv_polytopic}, due to $\mu$ and $\phi$. Also, the polytope \eqref{eq_ggv_polytopic} is expanded/contracted by the learnable function $S(a_z)$ in \eqref{eq_S_az}, which models the dependency of the maximum $a_y$ on the vertical accelerations $a_z$. Finally, \eqref{eq_ggv_comb_brake} defines a stability restriction of the cornering-braking performance, where $\bar{s}$ and $\bar{a}_{y}$ are learnable.
Section \ref{sec_learning_maximum_performance} will explain the learning process for the g-g-v constraints.
\subsection{E-NMPC Formulation} \label{sec_enmpc_formulation}
The KD vehicle model is integrated into the following minimum-time E-NMPC problem for trajectory planning:
\begin{equation}
	\label{eq_ENMPC}
	\begin{aligned}
	\min_{\boldsymbol{u} \in \mathcal{U}} \quad & \textstyle \int_{0}^T w_T \, \text{d} t + \sum_{j=1}^{7} \bigl(\boldsymbol{x}_j(T) - \boldsymbol{x}_{f_j} \bigr)^2\\
	\textrm{s.t.} \quad \textrm{KD model} \quad & \eqref{eq_MPC_mod_3D}, \eqref{eq_curvil_coords_1}, \eqref{eq_curvil_coords_2}, \eqref{eq_curvil_coords_3}\\
	\textrm{init. condit.} \quad & \boldsymbol{\mathcal{I}}(\boldsymbol{x}(0)) = 0 \\
	\textrm{constraints} \quad &
	\begin{cases}
		\eqref{eqn_proposed_ggv} \\
		-1 \leq \omega_{z_0} \leq 1 \\
		\mathrm{w} - d_r(\zeta) \leq n \leq d_l(\zeta) - \mathrm{w}
	\end{cases}
	\end{aligned}
\end{equation}
The cost function minimizes the maneuver time $T$, with $w_T$ being a tunable weight, and imposes soft final conditions for the states $\boldsymbol{x}$, as in \cite{Piccinini2023Access}. The optimization is subject to the KD vehicle model, the initial conditions $\boldsymbol{\mathcal{I}}(\boldsymbol{x}(0))$, the g-g-v constraints \eqref{eqn_proposed_ggv}, the bounds on the normalized yaw rate control $\omega_{z_0}$, and the track limits, expressed with the road width functions $d_r(\zeta)$ and $d_l(\zeta)$ (right and left margins) and the vehicle half-width $\mathrm{w}$.
Following \cite{Lot2014}, the OCP \eqref{eq_ENMPC} is reformulated using the curvilinear abscissa $\zeta$ as independent variable.
\subsection{Learning the Dynamics and Performance on 3D Circuits} \label{sec_learning_maximum_performance}
The related papers \cite{Rowold2023,Lovato2022,Lovato2022b,Massaro2020} assumed the prior knowledge of the vehicle simulator (VS) dynamic equations, and derived the maximum performance g-g-v diagram with QSS optimizations. In contrast, our ARD does not rely on the knowledge of the VS, and progressively learns the vehicle dynamics and the performance limits in 5 rounds. We extend the learning process in \cite{PiccininiARD2024} to 3D circuits:

$1^{\mathrm{st}}$ \textit{Round:} Open-loop steering-acceleration maneuvers (following \cite{Piccinini2021}) with the VS, on a 2D road, to preliminarily estimate the g-g-v diagram (\ref{eq_ax_min_max_2D},\ref{eq_ggv_polytopic}), and train the low-level tracking controllers (Fig. \ref{fig_planning_control}).

\begin{figure}[]
	\centering
	\includegraphics[width=0.8\columnwidth]{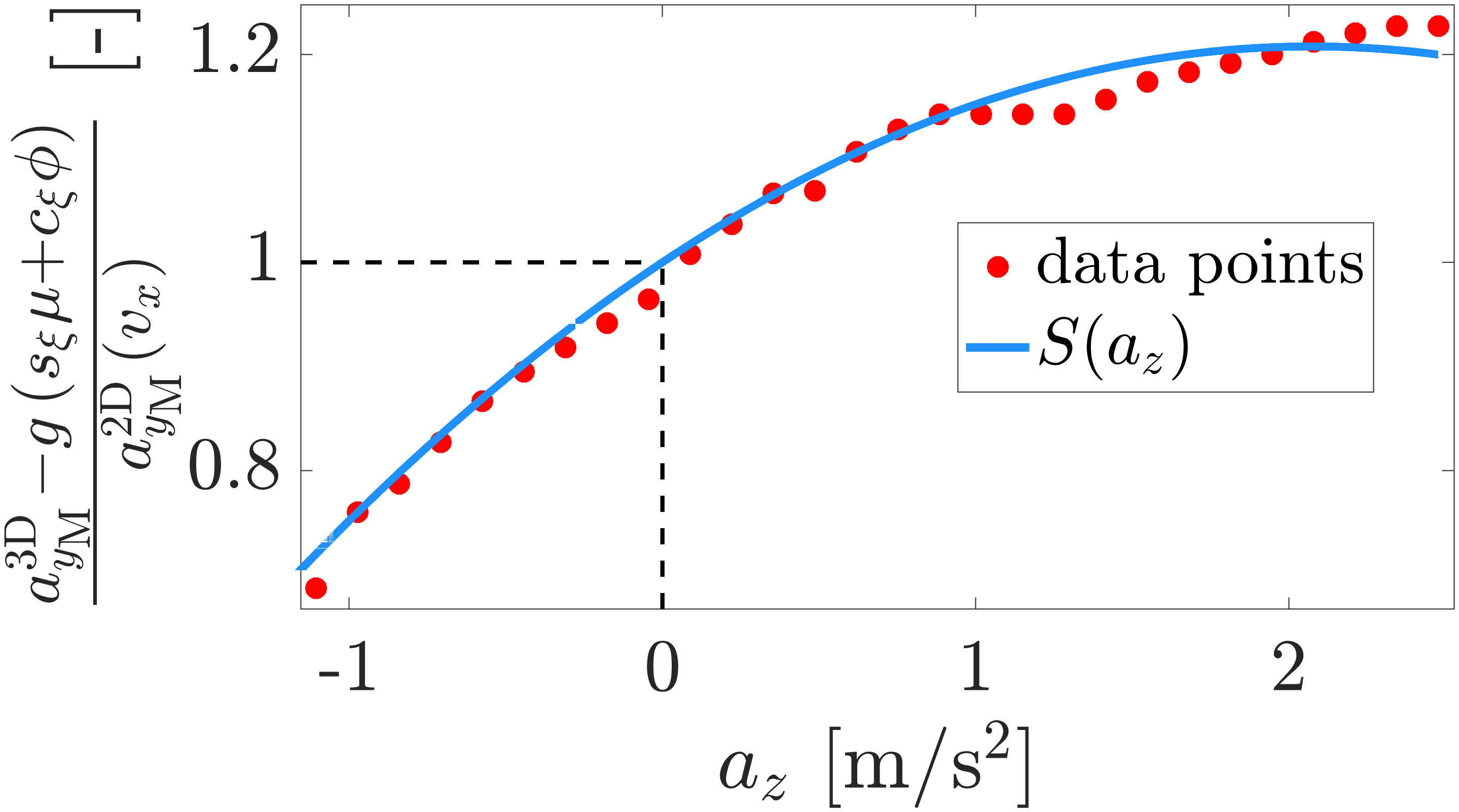}
	\caption{$S(a_z)$ polynomial function \eqref{eq_S_az}, to capture the deviations (red dots) between the measured peak lateral accelerations on a 3D circuit $a_{y_{\mathrm{M}}}^{\mathrm{3D}}$ and the 2D model $a_{y_{\mathrm{M}}}^{\mathrm{2D}}(v_x)$ in \eqref{eq_ayM_fcn_2D}. In the $5^{\mathrm{th}}$ learning round, ARD refines $S(a_z)$ with an online iterative learning method.}
	\label{fig_S_az_learned}
\end{figure}

$2^{\mathrm{nd}}$-$4^{\mathrm{th}}$ \textit{Rounds:} Closed-loop planning and control on a 3D circuit, using the previous-round controllers and g-g-v constraints \eqref{eqn_proposed_ggv}. Using the recorded vehicle telemetries, ARD iteratively refines the g-g-v constraints and low-level feedback controllers to minimize the lap time, as in \cite{PiccininiARD2024}. Also, ARD learns the 3D road-induced dynamics and performance: using the estimated vertical accelerations\footnote{$a_z$ is computed with \eqref{eq_az_omega_y}, and does not need to be measured.} $a_z$, it learns the lateral velocity model \eqref{eq_Fv_3D} and the $S(a_z)$ function \eqref{eq_S_az}. As shown in Fig. \ref{fig_ggv_2D_3D_top_Mugello_3D}, the maximum lateral accelerations on a 3D circuit significantly differ from the 2D case: $S(a_z)$ fits the deviations between $a_{y_{\mathrm{M}}}^{\mathrm{3D}}$ and $a_{y_{\mathrm{M}}}^{\mathrm{2D}}(v_x)$, being respectively the measured maximum lateral accelerations on a 3D circuit, and the 2D model in \eqref{eq_ayM_fcn_2D}. Fig. \ref{fig_S_az_learned} plots the fitted $S(a_z)$.

$5^{\mathrm{th}}$ \textit{Round:}
The $S(a_z)$ function \eqref{eq_S_az} heavily affects the lap times: overestimating it leads to infeasible lateral accelerations, while underestimating it yields suboptimal results. In the previous rounds, ARD learned $S(a_z)$ from closed-loop trajectories that satisfied the g-g-v constraints, and could not explore beyond the g-g-v limits.

To push beyond these constraints and decrease the lap times, the $5^{\mathrm{th}}$ round refines $S(a_z)$ with the iterative Nelder-Mead method \cite{Lagarias1998}. Each iteration consists of one lap, with the algorithm updating $S(a_z)$ until the lap time stops improving. This enables model-free online learning.
\section{Results} \label{sec_results}
\subsection{Vehicle Simulator, Racetrack and Hardware} \label{sec_results_settings}
In this section, our artificial race driver uses the proposed KD model for trajectory planning, and controls a high-fidelity double-track vehicle simulator (VS). Our VS is a high-fidelity model of a sports car, with a top speed of $320$ km/h. In \cite{PiccininiARD2024}, we validated the same VS with real telemetry data from a professional race driver.

We use the Mugello circuit (Italy), shown in Fig. \ref{fig_path_Mugello_3D}. The track geometry can be obtained from online tools like Google Earth Pro or from \cite{Leonelli2020,Marconi2020,Lovato2022}.

Our ARD is implemented in \texttt{C++} code. The software Pins \cite{Biral2016,Pagot2023Parking} is used to solve the E-NMPC problems in real-time, every 100 ms (10 Hz rate), with a prediction horizon of 300 m and 350 mesh points.
The results in this paper are obtained on a 2.6 GHz 6-Core Intel i7 processor.
\subsection{Analysis of the Negligible Terms} \label{sec_negligible_terms}
This section analyses which road-related terms in the KD model can be neglected, to decrease the computational times of the E-NMPC.
First, the expression of the vertical acceleration $a_z$ appearing in \eqref{eq_MPC_mod_3D_Omega}, \eqref{eq_MPC_mod_3D_vy}, \eqref{eq_ggv_polytopic}, \eqref{eq_ggv_comb_brake} can be simplified as follows:
\begin{subnumcases}{\label{eq_az_omega_y}}
	\omega_y = v_x \big(\xi \tau(\zeta) - \upsilon(\zeta)\big) \, / \, \big(1 - n \, \kappa(\zeta)\big) \label{eq_omega_y_simplif} \\
	a_z = - \omega_y v_x \label{eq_az_simplif}
\end{subnumcases}
where \eqref{eq_omega_y_simplif} and \eqref{eq_az_simplif} are obtained from respectively \eqref{eq_omega_y} and \eqref{eq_az_CoM_3D}, with a small-angle assumption for $\xi$ and neglecting $\dot{v}_z$. Finally, the 3D road-induced lateral and longitudinal accelerations in \eqref{eq_ayM_fcn_3D}, \eqref{eq_ax_min}, \eqref{eq_ax_max} are derived from \eqref{eq_ax_CoM_3D}, \eqref{eq_ay_CoM_3D}, neglecting the terms involving $\omega_y$ and $\omega_x$. Table \ref{tab_ARD_cpu_times} shows the impact of our reduced-complexity model: in comparison with the full KD model,
the computational times of E-NMPC decrease by a factor 2, while the lap time $T_{\mathrm{ARD}}$ with online closed-loop planning and control remains almost the same.
\begin{table}[]
	\centering
	\begin{tabular}{c|c|c|c}
		\toprule
		3D road terms & \multicolumn{2}{c|}{CPU time E-NMPC} & Lap time $T_{\mathrm{ARD}}$\\
		in the KD model & Mean & Std dev & (\textit{online} ARD)\\
		\hline
		All terms & 80.2 ms & 48.3 ms & 115.284 s \\
		Our model & 42.3 ms & 19.9 ms & 115.322 s \\
		\toprule
	\end{tabular}
	\caption{Compared with the full KD model, our version halves the E-NMPC's CPU times, with a minor effect on the lap time.}
	\label{tab_ARD_cpu_times}
\end{table}
\subsection{Benchmarks} \label{sec_benchmarks}
We compare the results of ARD with these benchmarks:
\begin{itemize}
	\item A minimum-lap-time OCP formulated with the high-fidelity VS (\textit{MLT-VS}), solved offline. The MLT-VS is our best lap time estimate for the VS.
	\item A minimum-lap-time OCP formulated with our KD vehicle model (\textit{MLT-KD}), solved offline. The MLT-KD yields the \textit{theoretical} lap time for the KD model, but does not consider the trajectory feasibility for the VS.
	\item The point-mass model and g-g-v constraints of \cite{Rowold2023}, which are integrated into our ARD for trajectory planning with E-NMPC, in place of our KD model.
\end{itemize}
\subsection{Artificial Race Driver Results}
\begin{table}[]
	\centering
	\begin{tabular}{c|c|c|c}
		\toprule
		\begin{tabular}{@{}c@{}}Learning\\round n\textdegree{}\end{tabular} & \begin{tabular}{@{}c@{}}Lap time $T_{\mathrm{VS}}$\\(\textit{offline} MLT-VS)\end{tabular} & \begin{tabular}{@{}c@{}}Lap time $T_{\mathrm{ARD}}$\\(\textit{online} ARD)\end{tabular} & \begin{tabular}{@{}c@{}}$\Delta T =$\\$T_{\mathrm{ARD}} - T_{\mathrm{VS}}$\end{tabular}\\
		\hline
		2 & \multirow{4}{*}{114.598 s} & 124.432 s & 9.834 s \\
		3 & & 119.127 s & 4.529 s \\
		4 & & 115.806 s & 1.208 s \\
		5 & & \textbf{115.322} s & \textbf{0.724} s\\
		\toprule
	\end{tabular}
	\caption{Comparing the lap times of the MLT-VS OCP and of our ARD, in the learning rounds on the Mugello 3D circuit.}
	\label{tab_res_learn_rounds}
\end{table}
\begin{table}[]
	\centering
	\begin{tabular}{ccc}
		\toprule
		\begin{tabular}{@{}c@{}}Learning\\round n\textdegree{}\end{tabular} & \begin{tabular}{@{}c@{}}Lap time\\$T_{\mathrm{VS}}$\\(MLT-VS)\end{tabular} & \begin{tabular}{@{}c@{}}Lap time\\$T_{\mathrm{ARD}}$\\(\textbf{ours})\end{tabular}\\
		\toprule
		2 & \multirow{4}{*}{114.59 s} & 124.43 s\\
		3 & & 119.12 s \\
		4 & & 115.80 s \\
		5 & & \textbf{115.32} s\\
		\toprule
	\end{tabular}
	\caption{Comparing the lap times of the MLT-VS OCP and of our ARD, in the learning rounds on the Mugello 3D circuit.}
	\label{tab_res_learn_rounds_}
\end{table}
\begin{table}[]
	\centering
	\begin{tabular}{c|c|c|c|c|c}
		\toprule
		KD &
		Mean & Lap time & Lap time & Lap time & $\Delta T =$\\
		model & CPU time & $T_{\mathrm{VS}}$ & $T_{\mathrm{KD}}$ & $T_{\mathrm{ARD}}$ & $T_{\mathrm{ARD}}$ \\
		& E-NMPC & MLT-VS & MLT-KD & ARD & $-T_{\mathrm{VS}}$ \\
		\hline
		\cite{Rowold2023} & 39.2 ms & \multirow{2}{*}{114.598 s} & 111.949 s & 116.110 s & 1.512 s\\
		%
		Ours & 42.3 ms & & 114.388 s & \textbf{115.322} s & \textbf{0.724} s\\
		\toprule
	\end{tabular}
	\caption{Comparing the performance of our KD model and the benchmark \cite{Rowold2023}, on the Mugello 3D circuit.}
	\label{tab_compare_with_benchmark}
\end{table}
Let us evaluate our ARD when performing closed-loop planning and control with the VS, using the settings of Section \ref{sec_results_settings}. A video with the main results is available as supplementary material.
\vspace{0.1cm}
\subsubsection{Comparison with MLT-VS OCP}
We now compare the performance of ARD and the MLT-VS OCP, on the Mugello circuit. Table \ref{tab_res_learn_rounds} reports the ARD's lap times in the learning rounds, compared with the MLT-VS. Since the MLT-VS uses the same VS model driven by ARD, then the MLT-VS's lap time is our best performance estimate.
The table shows a noticeable difference in the ARD's lap times over the learning rounds. In the $2^{\mathrm{nd}}$ round, ARD drives the vehicle with a limited knowledge of its dynamics and performance.
For this reason, ARD drives cautiously, and its lap time is $9.8$ s slower than the MLT-VS.
During the $3^{\mathrm{rd}}$ and $4^{\mathrm{th}}$ rounds, ARD learns the effect of the longitudinal and road-induced vertical accelerations on the vehicle dynamics: at the end of the $4^{\mathrm{th}}$ round, $T_{\mathrm{ARD}}$ is $1.2$ s slower than the MLT-VS.

Finally, in the $5^{\mathrm{th}}$ round, ARD refines its maximum performance estimates on the 3D circuit (Section \ref{sec_learning_maximum_performance}), and achieves a lap time $T_{\mathrm{ARD}} = 115.322$ s, which is $0.724$ s slower than the MLT-VS. This gap is mainly due to an imperfect learning of the performance gain induced by the 3D nature of the circuit. However, given the track length ($5.245$ km), we consider this difference acceptable for an autonomous agent. Indeed, the MLT-VS OCP represents an ideal lap time: it is achievable only without execution errors, and is typically hard to be matched by professional drivers, as our experience with telemetry data indicates.
\begin{figure}[]
	\centering
	\includegraphics[width=0.95\columnwidth]{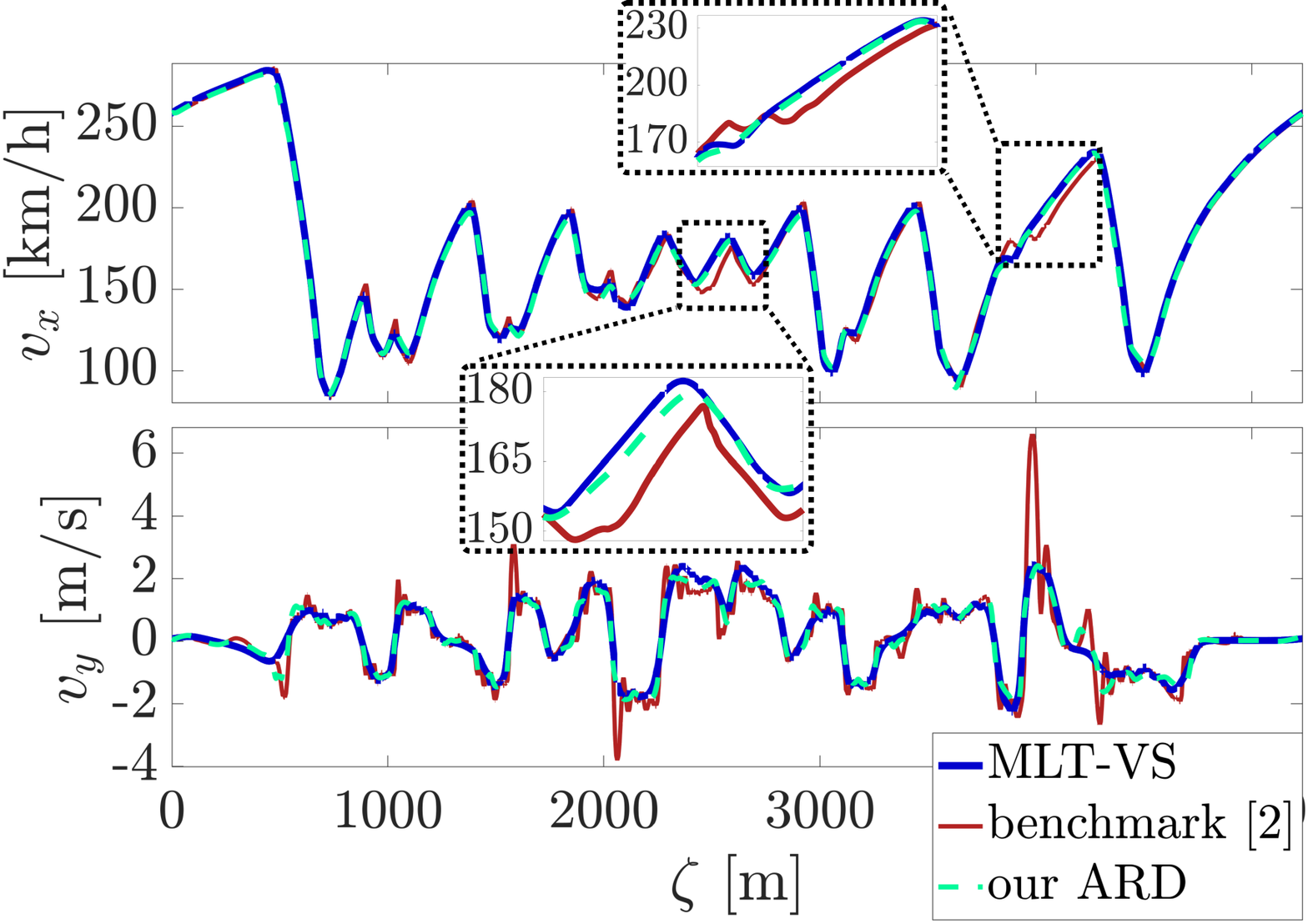}
	\caption{Comparing the longitudinal and lateral velocities of our ARD, the benchmark \cite{Rowold2023}, and the MLT-VS OCP, on the Mugello 3D circuit.}
	\label{fig_states_compare_Mugello_3D}
\end{figure}
\begin{figure}[]
	\centering
	\includegraphics[width=1\columnwidth]{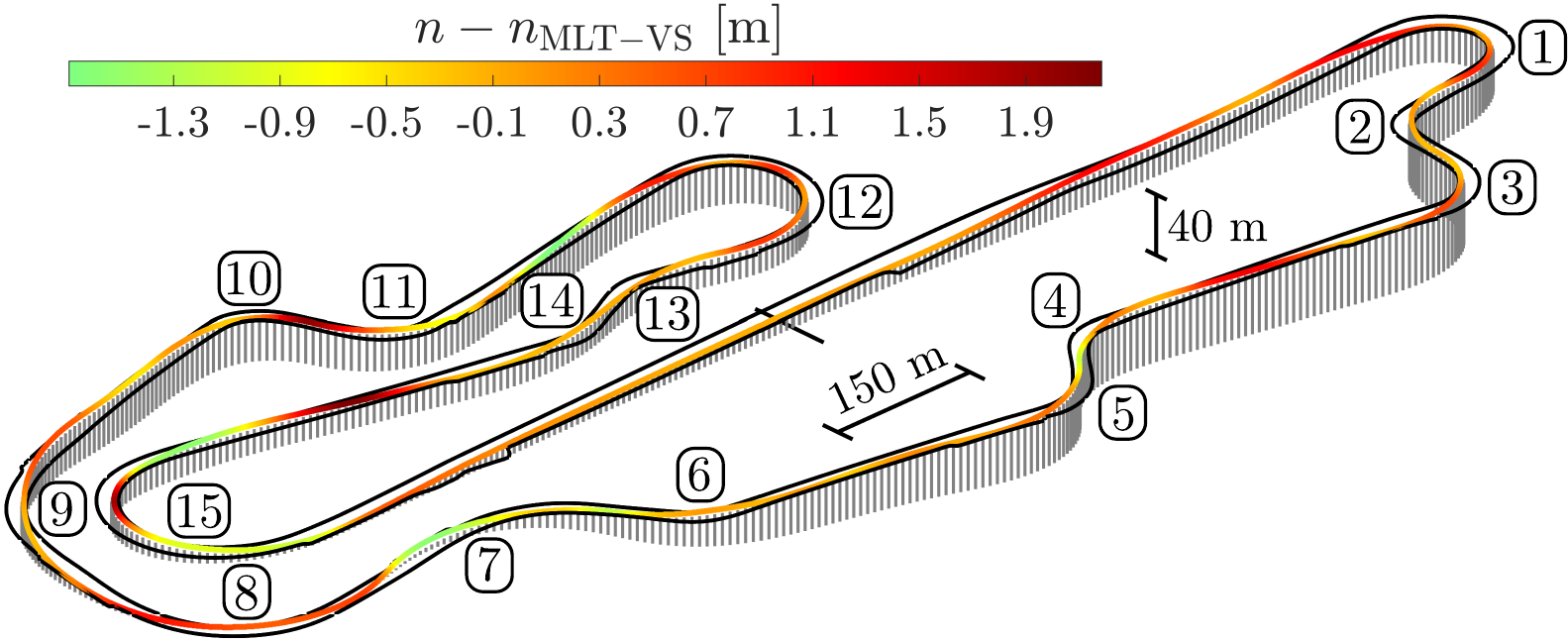}
	\caption{Trajectory executed by ARD on the Mugello 3D circuit. The color map represents the difference between the lateral coordinates of ARD and the MLT-VS OCP.}
	\label{fig_path_Mugello_3D}
\end{figure}
\begin{figure}[]
	\centering
	\includegraphics[width=1\columnwidth]{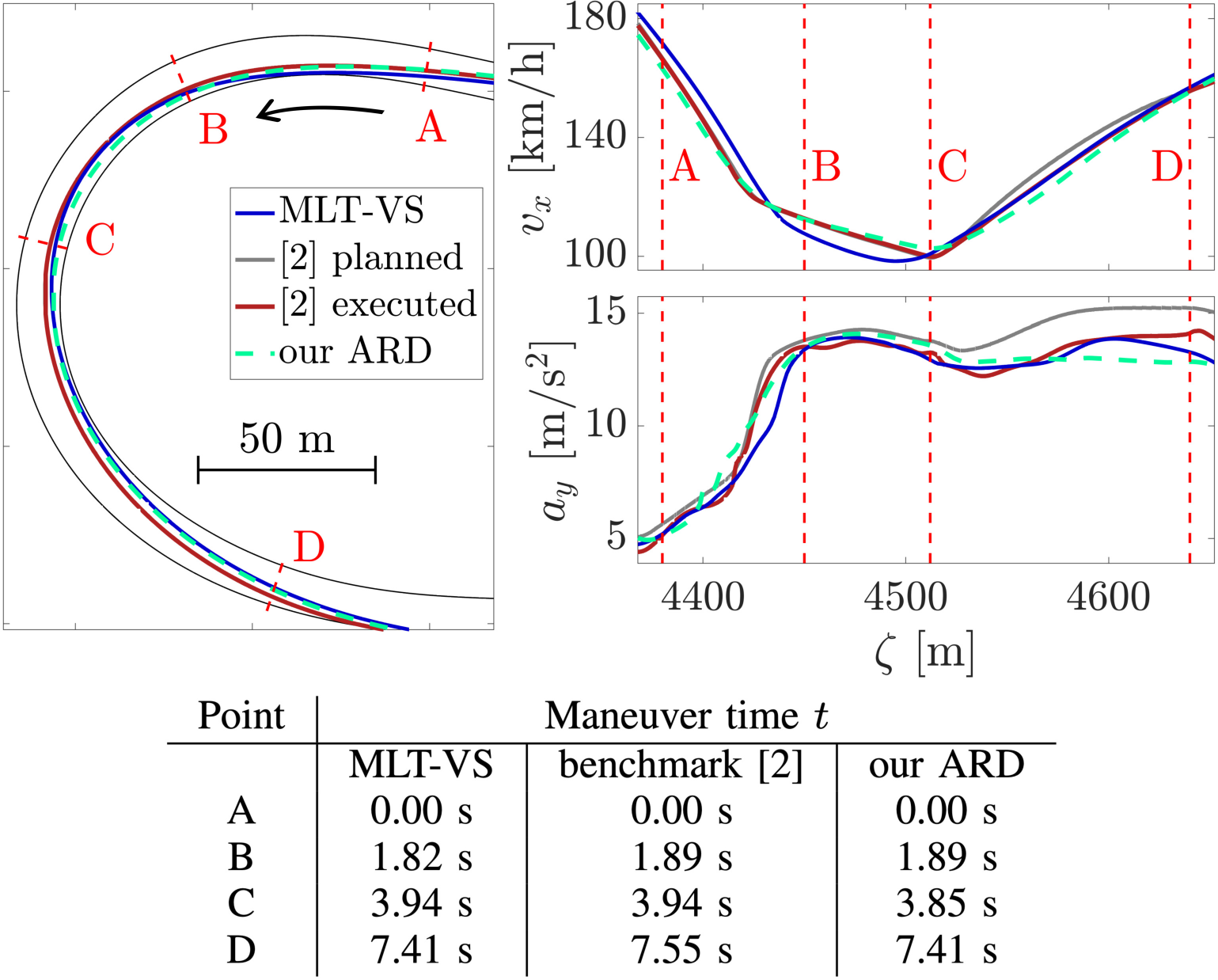}
	\caption{Comparing the trajectories and times of our ARD, the benchmark \cite{Rowold2023}, and the MLT-VS OCP, at the turn n.15 (\textit{Bucine}) of the Mugello circuit.}
	\label{fig_traject_compare_Mugello_3D}
\end{figure}
\subsubsection*{Trajectory Analyses}
Fig. \ref{fig_states_compare_Mugello_3D} shows that the longitudinal and lateral velocities executed by ARD are close to the MLT-VS solution. This indicates that ARD plans and executes feasible and near-optimal maneuvers, thanks to the new KD model for E-NMPC and the tracking controllers.
However, it is interesting to note how the executed trajectory locally deviates from the MLT-VS OCP, as shown in Fig. \ref{fig_path_Mugello_3D}. This is mainly due to two factors: (a) the learned performance and the KD planning model of ARD, which cannot perfectly match the actual VS, and (b) the ARD's capability to re-plan online new time-optimal maneuvers, to comply with local execution errors.

Fig. \ref{fig_traject_compare_Mugello_3D} analyzes the turn n.15. In the entry phase (point A), ARD is slower than the MLT-VS, and is on a wider path. In the middle of the corner (points B-C), ARD is faster and on a more internal path, but exits the turn with a lower speed (sector C-D). Interestingly, despite the different maneuvers, the A-D sector times are nearly identical: this happens with professional drivers as well \cite{Kegelman2017}.
\vspace{0.1cm}
\subsubsection{Comparison with MLT-KD OCP}
The MLT-KD OCP is solved offline, with the same KD model used by ARD for online planning. As shown in Table \ref{tab_compare_with_benchmark}, the MLT-KD lap time is $114.388$ s, which is close ($0.21$ s) to the MLT-VS. This means that the KD model and the g-g-v constraints are accurate estimates of the vehicle dynamics and performance. We stress that our ARD does \textit{not} track the MLT-KD trajectory, but rather re-plans and executes new time-optimal trajectories.
\vspace{0.1cm}
\subsubsection{Comparison with \cite{Rowold2023}} \label{sec_comparison_benchmark}
Let us now integrate the point-mass model and g-g-v constraints of \cite{Rowold2023} into the E-NMPC problem of our ARD. Compared with our KD model, the benchmark \cite{Rowold2023} differs in the following aspects:
\begin{itemize}
	\item Their g-g-v diagram is modeled with diamond shapes, which locally overestimate our real g-g-v (see \cite{Piccinini2024ggv,Rowold2023}).
	\item Their g-g-v diagram is estimated with QSS optimizations, assuming the knowledge of the VS's equations. In contrast, our ARD estimates the g-g-v while driving, with open and closed-loop maneuvers.
	\item Their point-mass model does not predict the lateral speed $v_y$, an important maneuver stability indicator \cite{DaLio2021}.
\end{itemize}
When the model and constraints of \cite{Rowold2023} are used in the offline MLT-KD OCP, the \textit{theoretical} lap time is $111.949$ s (Table \ref{tab_compare_with_benchmark}), which is $2.649$ s faster than the MLT-VS. However, this is due to an overestimation of the real g-g-v, and the resulting maneuvers are locally infeasible for the VS. Indeed, when the benchmark is integrated into ARD for closed-loop planning and control, the lap time is $116.110$ s, which is $1.512$ s slower than the MLT-VS, and $0.788$ s slower than ARD with our KD model. At the turn n.15 (Fig. \ref{fig_traject_compare_Mugello_3D}), the model of \cite{Rowold2023} plans lateral accelerations $a_y$ that are beyond the MLT-VS (sector B-D); however, the planned maneuver is infeasible, and ARD needs to locally re-plan and keep a wider path, thus losing time (0.14 s in sector A-D). Similar comments hold for other corners (Fig. \ref{fig_states_compare_Mugello_3D}), in which ARD reduces $v_x$ to comply with execution errors, or the vehicle slides laterally (peaks of $v_y$) due to an overestimated performance.
%
%
\subsection{Impact of the 3D Circuit Geometry} \label{sec_3D_road_geometry}
\begin{figure}[]
	\centering
	\includegraphics[width=1\columnwidth]{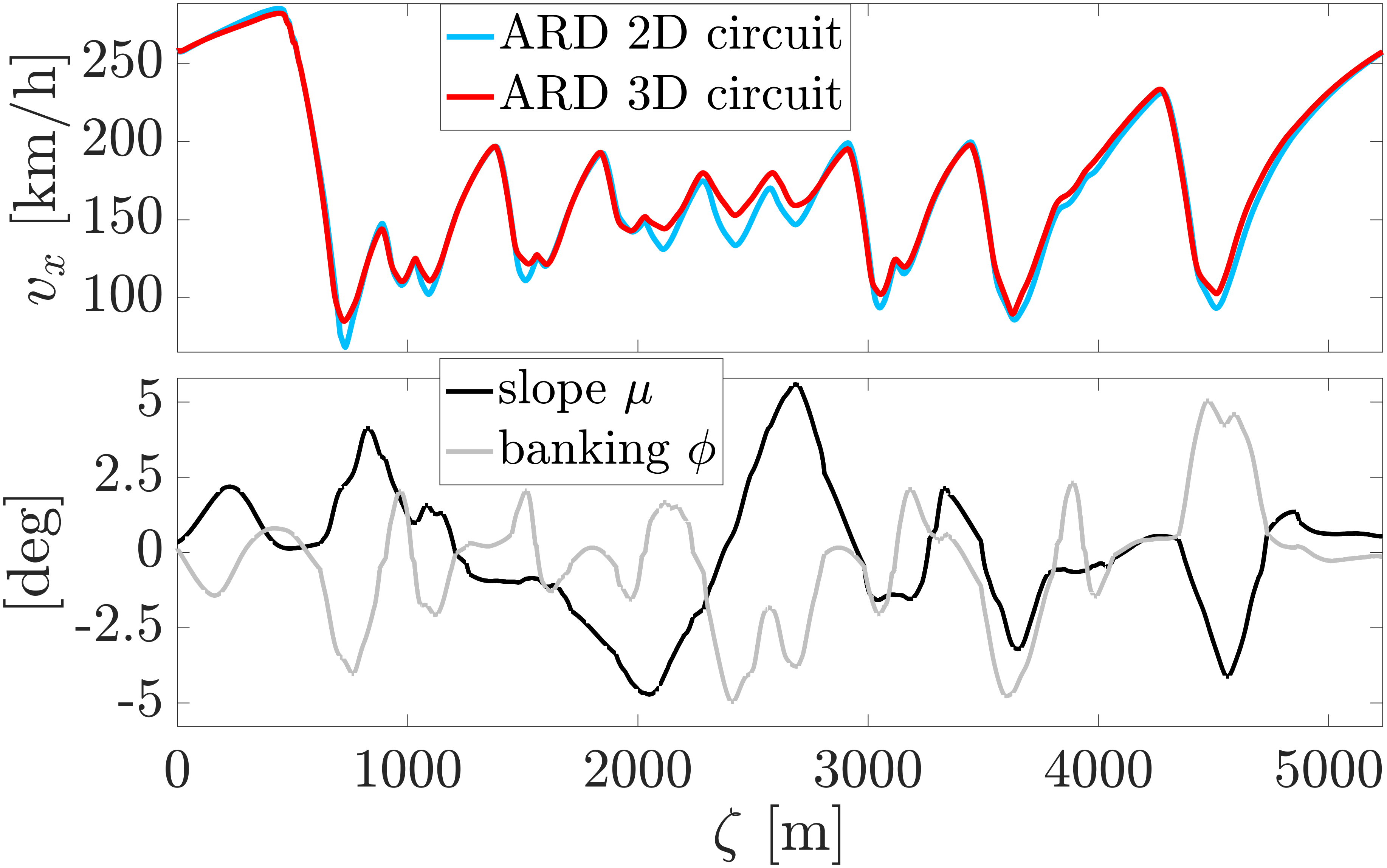}
	\caption{Comparing the speed profiles executed by ARD on the 2D and 3D versions of the Mugello circuit.}
	\label{fig_2D_vs_3D_Mugello_vx}
\end{figure}

\begin{table}[]
	\centering
	\begin{tabular}{c|c|c|c}
		\toprule
		\begin{tabular}{@{}c@{}}Circuit\\type\end{tabular} & \begin{tabular}{@{}c@{}}Lap time $T_{\mathrm{VS}}$\\(\textit{offline} MLT-VS)\end{tabular} & \begin{tabular}{@{}c@{}}Lap time $T_{\mathrm{ARD}}$\\(\textit{online} ARD)\end{tabular} & $T_{\mathrm{ARD}} - T_{\mathrm{VS}}$\\
		\hline
		2D & 118.741 s & 118.839 s & 0.098 s\\
		3D & 114.598 s & 115.322 s & 0.724 s\\
		\toprule
	\end{tabular}
	\caption{Comparing the lap times of the MLT-VS OCP and our ARD, on the 2D and 3D versions of the Mugello circuit.}
	\label{tab_2D_vs_3D}
\end{table}
Table \ref{tab_2D_vs_3D} compares the lap times on the Mugello 3D track and its artificial 2D version. In the 3D case, the MLT-VS's lap time is $4.1$ s faster: this difference is huge in Motorsport, and is confirmed by \cite{Lovato2022,Leonelli2020}. Our ARD decreases its lap time by $3.5$ s on the 3D circuit, by learning most of the road-induced performance gain. Fig. \ref{fig_2D_vs_3D_Mugello_vx} compares the speed $v_x$ profiles executed by ARD on the 2D and 3D racetrack versions. The $v_x$ difference reaches $20$ km/h in the corners n.8-9 ($\zeta \approx 2300 - 2700$ m), where the slope rates and the banking angles increase the vehicle performance (see the g-g-v plots in Fig. \ref{fig_ggv_2D_3D_top_Mugello_3D}). Also, on the 3D circuit, ARD is faster in many corner apexes, where it exploits the road banking to keep a higher speed.
\section{Conclusions and Future Work}  \label{sec:Conclusions}
This paper presented an artificial race driver (ARD) to plan and execute time-optimal maneuvers on 3D circuits. We developed a new kineto-dynamical (KD) model for trajectory planning with E-NMPC, and a learning process to estimate the vehicle dynamics and performance on 3D tracks. Our framework was tested with a high-fidelity vehicle simulator (VS) at the Mugello circuit. After learning the VS's dynamics, ARD achieved a lap time $0.724$ s slower than the MLT-VS OCP, which was our best performance estimate\textemdash a promising result for an autonomous agent. By analyzing the trajectories and lap times, our KD model was shown to outperform the benchmark \cite{Rowold2023} on the 3D circuit. Finally, the ARD's lap time was $3.5$ s faster on the 3D track compared to the artificial 2D version, which highlights the importance of learning the road-induced performance.

Future work will study the effect of 3D track curbs on the planned and executed maneuvers, by employing the physical tire model of \cite{Stocco2024} inside our VS. Also, we will integrate the proposed learning framework into a sampling-based motion planner for dynamic obstacle avoidance in autonomous racing \cite{Piazza2024}.
%
%
%
\bibliographystyle{IEEEtran}
\bibliography{IEEEabrv,biblio}
\end{document}